# Convergent message passing algorithms - a unifying view

Talya Meltzer, Amir Globerson and Yair Weiss
School of Computer Science and Engineering
The Hebrew University of Jerusalem, Jerusalem, Israel
{*talyam,gamir,yweiss*}*@cs.huji.ac.il*

## Abstract

Message-passing algorithms have emerged as powerful techniques for approximate inference in graphical models. When these algorithms converge, they can be shown to find local (or sometimes even global) optima of variational formulations to the inference problem. But many of the most popular algorithms are not guaranteed to converge. This has lead to recent interest in convergent message-passing algorithms.

In this paper, we present a unified view of convergent message-passing algorithms. We present a simple derivation of an abstract algorithm, tree-consistency bound optimization (TCBO) that is provably convergent in both its sum and max product forms. We then show that many of the existing convergent algorithms are instances of our TCBO algorithm, and obtain novel convergent algorithms "for free" by exchanging maximizations and summations in existing algorithms. In particular, we show that Wainwright's non-convergent sum-product algorithm for tree based variational bounds, is actually convergent with the right update order for the case where trees are monotonic chains.

## 1 Introduction

Probabilistic inference in graphical models is a key component in learning and using these models in practice. The two key inference problems are calculating the marginals of a model and calculating its most likely assignment (sometimes referred to as the MAP problem).

One approach to these generally intractable problems is to use a variational formulation, where approximate inference is cast as an optimization problem. For the marginals case this usually corresponds to minimization of a free energy functional, and for the MAP problem it corresponds to solving a linear programming (LP) relaxation [10].

A key challenge in both the MAP and marginals case is to devise simple and scalable algorithms for solving the variational optimization problem. In recent years numerous algorithms have been introduced for both tasks. These algorithms typically have a "message passing like" structure.

Perhaps the most widely used message-passing algorithms are "belief propagation" and its generalizations [5, 8, 12, 14]. These algorithms typically have two variants: sum-product which is used to approximate the marginals, and max-product which is used to approximate the MAP. Fixed-points of these algorithms can be shown to be local (or sometimes even global) optima of the corresponding variational formulation. Yet despite the spectacular empirical success of these algorithms in real-world applications, they are not guaranteed to converge, and variants of "dampening" are often used to improve their convergence [8, 12, 14].

There has therefore been much recent work on convergent message passing algorithms [2, 5, 7, 13]. These algorithms are often very similar in structure to the non-convergent algorithms and often include local max-product or sum-product operations. However, for each of these specific algorithms it has been possible to prove that the value of the variational problem (or its dual) improves at each iteration. Perhaps the most intriguing example of this is Kolmogorov's TRW-S algorithm [7] which is simply Wainwright's tree-reweighted max-product algorithm [11] with a different update schedule.

Here we introduce a unifying framework which encompasses both marginals and MAP approximations, by exploiting the mathematical similarities between these approximations. Specifically, we provide an up-per bound on the optimum of the variational approximations, and give sufficient conditions that algorithms need to satisfy in order to decrease this bound in a monotone fashion. Any algorithm which satisfies these conditions is guaranteed to decrease the upper bound at every iteration. This property in turn guarantees that such algorithms converge to a global optimum of the variational problem in the marginals case and to a local optimum in the MAP LP approximation case.

Our framework involves updating a subset of regions which form a tree in the region graph. A related approach was recently suggested by Sontag *et al.* [9] in the context of solving the MAP approximation. Their work gives an explicit algorithm for optimizing all edges corresponding to a tree in the original graph, such that an upper bound on the LP optimum is decreased at every iteration. Our formalism does not give an explicit update but rather conditions that guarantee an update to decrease the objective. However, we show that these conditions are satisfied by several known algorithms. Furthermore, since the condition is similar in the marginals and MAP case (specifically a condition of sum and max consistency respectively) it is easy to obtain algorithms for both these cases simultaneously, and to use results in one problem for obtaining algorithms for the other.

For instance, we consider the tree-reweighted (TRW) free energy minimization problem [12]. Recently two works have provided convergent algorithms for this problem [1, 3], but these were more involved than standard message passing algorithms. Here we show the surprising result that in fact the original algorithm provided for TRW by Wainwright *et al.* is convergent, if run with an appropriate schedule of message updates.

## 2 Bounds for MAP and Log-Partition

We consider a graphical model where the joint probability over variables $p(x)$ factors into a product over clique potentials $p(x) = \frac{1}{Z} \prod_\alpha \Psi_\alpha(x_\alpha)$ or equivalently, the energy function is a sum over clique energies $p(x) = \frac{1}{Z} \exp(\sum_\alpha \theta_\alpha(x_\alpha))$. We also denote $\theta(x) = \sum_\alpha \theta_\alpha(x_\alpha)$.

The problem of calculating marginals and approximation of the partition function $Z$ can be recast as the following maximization problem of the function $F(q)$ (the negative of the free energy):

$$\log Z = \max_q F(q) = \max_q \left( \langle \theta(x) \rangle_q + H(q) \right) \quad (1)$$

where $q$ is the set of probability distributions over $x$, $\langle \theta(x) \rangle_q$ is the average energy with respect to $q$ and $H(q)$ is the entropy function. The maximizing argument is then the distribution $p(x)$.

This maximization is in general intractable, so approximate free energies are often used. A class of approximate free energies, discussed in [14], is based on the concept of a region graph $G$ whose nodes $\alpha$ are regions of the original graph, and whose edges represent subregion relationships (i.e. an edge between $\alpha$ in $\beta$ exists only if $\beta \subset \alpha$). The approximation replaces the joint entropy $H(q)$ with a linear combination of local region entropies $H_\alpha(q_\alpha)$ where each local entropy is weighted by a "double-counting" number $c_\alpha$:

$$\tilde{H}_{G,c}(q) = \sum_\alpha c_\alpha H_\alpha(q_\alpha) , \quad (2)$$

where the subscript $G, c$ indicates the dependence of the approximation on the region graph $G$ and the counting numbers $c_\alpha$.

With this approximation of $H(q)$ the free energy now only depends on local distributions, since the average energy is a simple function of the $q_\alpha$, namely $\langle \theta(x) \rangle_q = \sum_\alpha \langle \theta_\alpha(x_\alpha) \rangle_{q_\alpha}$. To optimize only over local distributions $q_\alpha$, we need to consider only distributions such that there exists a $q(x)$ that has these as marginals. This set (called the marginal polytope [10]) cannot be expressed in a compact form, and is typically approximated. One popular approximation is the local polytope of the region graph $L(G)$ defined as the set of local distributions that agree on the marginals for any two regions in the region graph that are subsets of each other:[1]

$$L(G) = \left\{ q \geq 0 \ \middle| \ \begin{array}{l} \forall \beta \subset \alpha, x_\beta \sum_{x_\alpha \setminus x_\beta} q_\alpha(x_\alpha) = q_\beta(x_\beta) \\ \forall \alpha, \ \sum_{x_\alpha} q_\alpha(x_\alpha) = 1 \end{array} \right\}$$

Take together, this results in the following standard variational approximation [10]:

$$\max_{q \in L(G)} \tilde{F}(q) = \max_{q \in L(G)} \sum_\alpha \langle \theta_\alpha(x_\alpha) \rangle_{q_\alpha} + \tilde{H}_{G,c}(q) \quad (3)$$

Similarly the MAP problem is approximated via

$$\max_{q \in L(G)} \tilde{F}(q) = \max_{q \in L(G)} \sum_\alpha \langle \theta_\alpha(x_\alpha) \rangle_{q_\alpha} \quad (4)$$

To obtain a unified formalism for MAP and marginals, we use a temperature parameter $T$ where $T = 1$ for marginals and $T = 0$ for MAP and the optimization is:

$$\max_{q \in L(G)} \tilde{F}(q) = \max_{q \in L(G)} \sum_\alpha \langle \theta_\alpha(x_\alpha) \rangle_{q_\alpha} + T \tilde{H}_{G,c}(q) \quad (5)$$

---
[1] Note that this is the local polytope of the region graph *not* the local polytope of the original graph

### 2.1 Positive Counting Numbers

The entropy approximation $\tilde{H}_{G,c}(q)$ in Eq. 2 is generally not a concave function of the local distributions $q$. Thus maximization of $\tilde{F}(q)$ may result in local optima. To avoid this undesirable property, several works (e.g., [5]) have focused on entropies which are obtained by considering only concave $\tilde{H}_{G,c}(q)$ functions. We focus on approximations where all the double counting numbers are *non-negative*. This is a strong restriction but since we are working with a region graph formulation, many approximate free energies which have negative double counting numbers can be transformed into ones with positive double counting numbers on a region graph. Perhaps the most important example are *tree-reweighted* free energies in which the entropy is approximated as $H_{TRBP}(q) = \sum_\tau \rho_\tau H_\tau(q)$ with $\rho_\tau$ a probability distribution over trees in the graph and $H_\tau(q)$ is the entropy of the distribution on $\tau$ with marginals given by $q$ (more precisely, the projection of $q$ on the tree $\tau$). If we consider a region graph with trees and their intersection (Fig. 5) the double counting numbers are non-negative. But $H_{TRBP}$ can also be rewritten $H_{TRBP} = \sum_{ij} \rho_{ij} H_{ij} + \sum_i c_i H_i$ with $c_i = 1 - \sum_j \rho_{ij}$ and $\rho_{ij}$ is the edge appearance probability of the edge $ij$ under the distribution $\rho$. In this case, the double-counting number for the singletons $c_i$ may be negative. However, we will show that it is sometimes advantageous to work in the representation that uses a non-negative mixture of trees, since nonnegativity of the counting numbers allows a simpler derivation of algorithms.

### 2.2 Optimization and Reparameterization

The vast majority of methods for solving the variational approximation are based on two classes of constraints that local optima should satisfy. It is easy to show using Lagrange multipliers, that local optima of $\tilde{F}$ should satisfy two types of constraints [5, 6, 12, 14, 15]

- Reparametrization (or admissibility, or "e constraints"). $P(x) \propto \prod_\alpha q_\alpha(x_\alpha)^{c_\alpha}$, for every $x$.
- sum-consistency (or "m constraints"), $\sum_{x_\alpha \setminus x_\beta} q_\alpha(x_\alpha) = q_\beta(x_\beta)$ for all $\beta \subset \alpha$ and $x_\beta$.

By enforcing each of these constraints iteratively one obtains many of the popular sum-product algorithms. Replacing the sum-consistency constraint with max-consistency gives many of the popular max-product algorithms. A simple example is ordinary BP, which maintains admissiblity at each iteration and a message from $i$ to $j$ enforces consistency between $b_{ij}$ and $b_j$.

In general, simply iteratively enforcing constraints is not guaranteed to give convergent algorithms. However, as we show in this paper, by iterating through the constraints in a particular order, we obtain monotonically convergent algorithms.

### 2.3 Bound minimization and reparameterizations

We begin by providing an upper bound on the log-partition function whose minimization is equivalent to the maximization in Eq. 5.

We consider marginals $b_\alpha$ of the exponential form:

$$b_\alpha(x_\alpha; \tilde{\theta}_\alpha) = \frac{1}{Z_{\tilde{\theta}_\alpha}} \exp\left(\tilde{\theta}_\alpha(x_\alpha)/c_\alpha\right) \quad (6)$$

and require that these marginals will be admissible (maintain the "e constraints"). We obtain admissibility by requiring that the variables $\tilde{\theta}$ will satisfy the following for each $x$:

$$\sum_\alpha \theta_\alpha(x_\alpha) = \sum_\alpha \tilde{\theta}_\alpha(x_\alpha) \quad (7)$$

The algorithms we propose will optimize over the variables $\tilde{\theta}$ while keeping the constraint in Eq. 7 satisfied at all times. Moreover, they will monotonically decrease an upper bound on the optimum of Eq. 5. In the following two lemmas we provide this bound for the sum and max cases.

**Lemma 1** *The approximation to the log partition function is bounded above by:*

$$bound_{sum}(\tilde{\theta}) = \sum_\alpha c_\alpha \ln Z_{\tilde{\theta}_\alpha} \quad (8)$$

*for any reparameterization $\tilde{\theta}$ (i.e., any $\tilde{\theta}$ satisfying Eq. 7).*

*Minimizing $bound_{sum}(\tilde{\theta})$ over the set of reparameterizations $\tilde{\theta}$ would give the approximated log-partition function:*

$$\min_{\tilde{\theta}} bound_{sum}(\tilde{\theta}) = \max_{q \in L(G)} \tilde{F}(q) \quad (9)$$

*This is the optimum of Eq. 5 with $T = 1$.*

**Proof:** Kolmogorov [7] showed that if $\tilde{\theta}$ is a reparameterization (i.e. keeping the constraint in Eq. 7), it also holds that $\langle \tilde{\theta} \rangle_q = \langle \theta \rangle_q$ for any $q \in L(G)$. Using this property, we can see that the log-partition function is constant under reparameterization: $\tilde{F}(q; \theta) = \tilde{F}(q; \tilde{\theta})$ for any $q \in L(G)$, and in particular the maximum value will remain the same. Now, using the new variables $\tilde{\theta}$ we have a trivial bound on the log-partition:

$$\max_{q \in L(G)} \tilde{F}(q;\tilde{\theta}) = \max_{q \in L(G)} \sum_\alpha \left( \langle \tilde{\theta}_\alpha \rangle_{q_\alpha} + c_\alpha H_\alpha(q_\alpha) \right)$$
$$\leq \sum_\alpha \max_{q_\alpha} \left( \langle \tilde{\theta}_\alpha \rangle_{q_\alpha} + c_\alpha H_\alpha(q_\alpha) \right)$$

Since the counting numbers $c_\alpha$ are non-negative, the marginals defined in Eq. 6 maximize each local functional $F_\alpha(q_\alpha; \tilde{\theta}_\alpha, c_\alpha) = \langle \tilde{\theta}_\alpha \rangle_{q_\alpha} + c_\alpha H_\alpha(q_\alpha)$, and the optimal value is $c_\alpha \ln Z_{\tilde{\theta}_\alpha}$. Thus,

$$\max_{q \in L(G)} \tilde{F}(q;\tilde{\theta}) \leq \sum_\alpha c_\alpha \ln Z_{\tilde{\theta}_\alpha} \quad (10)$$

The bound is tight if there exists a reparameterization $\tilde{\theta}$ such that the marginals $b_\alpha(x_\alpha; \tilde{\theta}_\alpha)$ are sum-consistent (i.e. $b \in L(G)$). The existence of such a re-parameterization is guaranteed if the maximum of the approximated negative free energy $\tilde{F}(q;\theta)$ does not happen at an extreme point [14].

A similar result may be obtained for the MAP case (this result or variants of it appeared in previous works, e.g., [7, 9, 13]).

**Lemma 2** *The value of the MAP is bounded above by:*

$$bound_{max}(\tilde{\theta}) = \sum_\alpha \max_{x_\alpha} \tilde{\theta}_\alpha(x_\alpha) \quad (11)$$

*for any reparameterization $\tilde{\theta}$ (i.e., any $\tilde{\theta}$ satisfying Eq. 7).*

*Minimizing $bound_{max}(\tilde{\theta})$ over the set of reparameterizations $\tilde{\theta}$ would give the optimal value for the region-graph LP relaxation of the MAP:*

$$\min_{\tilde{\theta}} bound_{max}(\tilde{\theta}) = \max_{q \in L(G)} \sum_\alpha \langle \theta_\alpha(x_\alpha) \rangle_{q_\alpha} \quad (12)$$

*This is the optimum of Eq. 5 with $T = 0$.*

**Proof:** The bound follows directly from the admissiblity constraint so that $\max_x \sum_\alpha \tilde{\theta}_\alpha(x_\alpha) \leq \sum_\alpha \max_{x_\alpha} \tilde{\theta}_\alpha(x_\alpha)$. The fact that the tightest bound coincides with the LP relaxation was proven in [13].

We note that the above two lemmas may also be viewed as an outcome of convex duality. In other words, the original variational maximization problem and the equivalent bound minimization problem are convex duals of each other.

In the following sections we provide a framework for deriving minimization algorithms for the above two bounds.

---

**Algorithm 1** The tree consistency bound optimization (TCBO) algorithm

Iterate over sub-graphs $T$ of the region graph that have a tree structure:

1. Choose a tree $T$
2. Update the values of $\tilde{\theta}_\alpha^{t+1}$ for all $\alpha \in T$ such that:

   - *re-parameterization* is maintained:
     $$\theta(x) = \sum_{\alpha \in T} \tilde{\theta}_\alpha^{t+1}(x_\alpha) + \sum_{\alpha \notin T} \tilde{\theta}_\alpha^t(x_\alpha)$$

   - *tree-consistency* is enforced:
     Define the beliefs
     $$b_\alpha^{t+1}(x_\alpha; \tilde{\theta}_\alpha^{t+1}) = \frac{1}{Z_\alpha^{t+1}} \exp\left( \tilde{\theta}_\alpha^{t+1}(x_\alpha)/c_\alpha \right)$$

     For each $\alpha \in T, \beta \in T, \beta \subset \alpha$ and $x_\beta$, optimize the bound to the log-partition function by enforcing *sum-consistency*:
     $$\sum_{x_{\alpha \setminus \beta}} b_\alpha^{t+1}(x_\alpha; \tilde{\theta}_\alpha^{t+1}) = b_\beta^{t+1}(x_\beta; \tilde{\theta}_\beta^{t+1})$$

     or optimize the bound to the MAP by enforcing *max-consistency*:
     $$\max_{x_{\alpha \setminus \beta}} b_\alpha^{t+1}(x_\alpha; \tilde{\theta}_\alpha^{t+1}) = b_\beta^{t+1}(x_\beta; \tilde{\theta}_\beta^{t+1})$$

---

## 3 Bound optimization and consistency

We propose the tree consistency bound optimization (TCBO) algorithm as a general framework for minimizing the bounds in Sec. 2.3 for the approximated log-partition and for the MAP, within a region graph with positive counting numbers $c_\alpha$.

The idea is to perform updates on trees that are subgraphs of the region-graph. The $\tilde{\theta}$ corresponding to each such tree will be updated simultaneously in a way that will achieve a monotone decrease in the bound.

The method we propose, as described in Algorithm 1 keeps the beliefs admissible with the positive counting numbers $c_\alpha$ (or equivalently, always maintains $\tilde{\theta}(x)$ that reparameterize the original energy $\theta(x)$). The corresponding $\tilde{\theta}$ thus satisfy the conditions of the bound in Sec. 2.3. Furthermore, at each iteration, max or sum consistency of the beliefs is enforced for the subtree $T$.

As mentioned earlier, maintaining consistency on subsets does not generally result in convergent algorithms. However, as the following lemmas show, in our case enforcing consistency is equivalent to block coordinate descent on the bound.

**Lemma 3 The sum-consistency lemma:** *Consider the bound minimization problem for the log-partition function with positive counting numbers (Lemma 1), defined on a subset of regions and intersections $T$. The part of the bound which is influenced by the beliefs of the subset is:*

$$PB_T(\tilde{\theta}_T) = \sum_{\alpha \in T} c_\alpha \ln Z_{\tilde{\theta}_\alpha}$$

*The problem is to find $\{\tilde{\theta}_\alpha\}$ for all $\alpha \in T$ that minimize $PB_T(\tilde{\theta}_T)$ subject to $\tilde{\theta}$ being reparameterizations of the energy $\{\theta_\alpha\}$. If for some reparameterization $\tilde{\theta}^*$ the beliefs $b(x_\alpha; \tilde{\theta}^*_\alpha) \propto \exp\left(\tilde{\theta}^*_\alpha/c_\alpha\right)$ are sum-consistent, then it minimizes the bound.*

**Proof:** The part of the bound which is dependent on $\tilde{\theta}_T$ the is bounded below:

$$\begin{aligned} PB_T(\tilde{\theta}_T) &= \sum_{\alpha \in T} \max_{q_\alpha} F_\alpha(q_\alpha; \tilde{\theta}_\alpha, c_\alpha) \\ &\geq \max_{q_T \in L(G)} \sum_{\alpha \in T} F_\alpha(q_\alpha; \tilde{\theta}_\alpha, c_\alpha) \end{aligned}$$

Now, if we find variables $\tilde{\theta}_\alpha$ for all $\alpha \in T$ such that they provide global re-parameterization $\tilde{\theta}(x) = \theta(x)$ (so we can have a bound), and the marginals $b_\alpha(x_\alpha; \tilde{\theta}_\alpha) \propto \exp(\tilde{\theta}_\alpha(x_\alpha)/c_\alpha)$ which maximize each term $F_\alpha(q_\alpha; \tilde{\theta}_\alpha, c_\alpha)$ separately are also sum-consistent ($b_T \in L(G)$), then $PB_T(\tilde{\theta}_T)$ will achieve its optimal value, and thus we perform block coordinate descent on the bound. □

Note that for optimizing the bound to the log-partition, the subset $T$ does not have to form a tree, and the sum-consistency of the beliefs is enough. Yet, it may be easier in practice to enforce sum-consistency on trees.

**Lemma 4 The max-consistency lemma:** *Consider the bound minimization problem for MAP with positive counting numbers (Lemma 2), defined on a subset of regions and intersections that form a tree $T$. The part of the bound which is influenced by the beliefs of the tree is:*

$$PB_T(\tilde{\theta}_T) = \sum_{\alpha \in T} \max_{x_\alpha} \tilde{\theta}_\alpha(x_\alpha)$$

*The problem is to find $\{\tilde{\theta}_\alpha\}$ for all $\alpha \in T$ that minimize $PB_T(\tilde{\theta}_T)$ subject to $\tilde{\theta}$ being reparameterizations of the energy $\{\theta_\alpha\}$. If for some reparameterization $\tilde{\theta}^*$ the beliefs $b(x_\alpha; \tilde{\theta}^*_\alpha) \propto \exp\left(\tilde{\theta}^*_\alpha/c_\alpha\right)$ are max-consistent, then it minimizes the bound.*

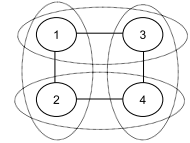

Figure 1: A simple 2x2 grid, with pair regions.

**Proof:** The part of the bound which is dependent on $\tilde{\theta}_T$ is bounded below:

$$PB_T(\tilde{\theta}_T) \geq \max_{x_T} \tilde{\theta}_T(x_T)$$

where

$$\tilde{\theta}_T(x_T) \doteq \sum_{\alpha \in T} \tilde{\theta}_\alpha(x_\alpha)$$

so if we can find an assignment $x^*_T$ whose cost $\tilde{\theta}_T(x^*_T)$ equals $PB_T(\tilde{\theta}_T)$, that means we have the tightest bound. Now, if for some reparameterization $\tilde{\theta}^*_T$ the beliefs $b_\alpha(x_\alpha; \tilde{\theta}^*_\alpha) \propto \exp\left(\tilde{\theta}^*_\alpha/c_\alpha\right)$ are max-marginalizable then we can always find an assignment $x^*_T$ that sits on the maxima of $\tilde{\theta}^*_\alpha$ because the subgraph is a tree (so there cannot be any frustrations). Hence, we obtain $\tilde{\theta}_T(x^*_T) = PB_T(\tilde{\theta}^*_T)$, and the bound achieves its optimal value for the coordinates in $T$. □

The above two lemmas show that the TCBO algorithm monotonically decreases the bound after each update. In the log-partition case, the bound is strictly convex and thus this strategy finds the global minimum of the bound which is the global maximum of Eq. 5. In the MAP case, the function is not strictly convex and the algorithm may converge to values that are not its global optimum. This phenomenon is shared by most dual descent algorithms (e.g., [2, 7, 13]).

TCBO is a general scheme and can be implemented for different choices of tree sub-graphs. In the next section we illustrate some possible choices and their relation to known algorithms.

## 4 Existing bound minimizing algorithms

We identify some existing convergent algorithms as instances of TCBO: Heskes' algorithm [5] for approximating the marginals, and MPLP [2], TRW-S [7], max-sum diffusion (MSD) [13] for approximating MAP.

Figures 2-5 show the reparametrization, region graph and the tree sub-graph updated at each iteration of these algorithms, for the simple example of a 2x2 grid shown in Fig. 1. Note that all algorithms use a reparameterization with positive double counting numbers. Furthermore, they update only a subtree at

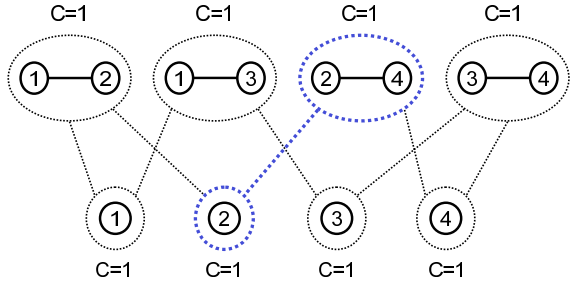

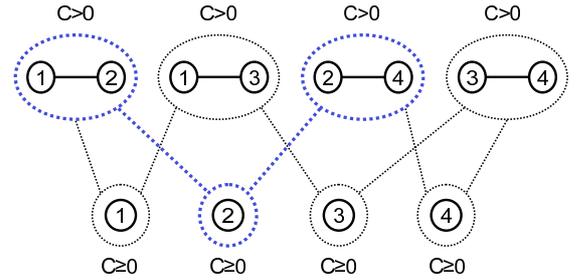

Figure 2: Illustration of the max-sum-diffusion (MSD) algorithm as an instance of the TCBO formalism. MSD operates on a region graph containing pairs and singletons (here corresponding to a 2x2 grid). The subgraph T corresponding to the TCBO update is shown in the blue dashed line.

Figure 3: Heskes' sum-product algorithm may be viewed as a TCBO algorithm updating the subtree shown in the blue dashed line (a star graph centered on a singleton node).

**Algorithm 2** The max sum diffusion (MSD) algorithm

Iterate over edges between regions $<\alpha, \beta>$:

1. Set the message from $\alpha$ to $\beta$:

$$m_{\alpha \to \beta}^{t+1}(x_\beta) = m_{\alpha \to \beta}^t(x_\beta) \cdot \sqrt{\frac{\max_{x_{\alpha \setminus \beta}} b_\alpha^t(x_\alpha)}{b_\beta^t(x_\beta)}}$$

2. Update the beliefs:

$$b_\beta^{t+1}(x_\beta) \propto m_{\alpha \to \beta}^{t+1}(x_\beta) \cdot \prod_{\alpha' \neq \alpha} m_{\alpha' \to \beta}^t(x_\beta)$$

$$b_\alpha^{t+1}(x_\alpha) \propto \frac{\Psi_\alpha(x_\alpha)}{m_{\alpha \to \beta}^{t+1}(x_\beta) \cdot \prod_{\beta' \neq \beta} m_{\alpha \to \beta'}^t(x_\beta)}$$

**Algorithm 3** Heskes' sum-product algorithm

Iterate over intersection regions $\beta$:

1. $\forall \alpha \supset \beta$ set the message from $\alpha$ to $\beta$:

$$m_{\alpha \to \beta}^{t+1}(x_\beta) = \frac{\sum_{x_{\alpha \setminus \beta}} b_\alpha^t(x_\alpha)}{m_{\beta \to \alpha}^t(x_\beta)}$$

2. Update the belief of the intersection region:

$$b_\beta^{t+1}(x_\beta) \propto \prod_{\alpha \supset \beta} \left( m_{\alpha \to \beta}^{t+1}(x_\beta) \right)^{c_\alpha / \hat{c}_\beta}$$

3. $\forall \alpha \supset \beta$ set the messages to the parent regions and their beliefs:

$$m_{\beta \to \alpha}^{t+1}(x_\beta) = \frac{b_\beta^{t+1}(x_\beta)}{m_{\alpha \to \beta}^{t+1}(x_\beta)}$$

$$b_\alpha^{t+1}(x_\alpha) \propto \Psi_\alpha^{1/c_\alpha}(x_\alpha) \cdot m_{\beta \to \alpha}^{t+1}(x_\alpha) \prod_{\beta' \neq \beta} m_{\beta' \to \alpha}^t(x_\beta')$$

a time. What remains to be shown is that each iteration achieves consistency among the beliefs (in other words, it satisfies the conditions of TCBO framework and thus monotonically decreases the corresponding upper bound).

Heskes' algorithm can be shown to be an instance of TCBO using direct substitution. The update rules are shown in algorithm 3. MPLP (algorithm 4) does not appear at first sight to use the region graph illustrated in Fig. 4, but rather works with edges and singletons. However, as we show in the appendix, there is a way to transform the messages used in the max-product version of Heskes's algorithm into messages of MPLP using the MPLP region graph. The max-consistency achieved by MSD (algorithm 2) can again be shown directly.

It is also possible to use tree graphs (or forests) as regions, and various existing methods indeed use this approach. We may consider a TCBO algorithm which iterates through all edges and nodes, and for each edge or node enforces consistency between it and all trees that contain it. This is illustrated in Fig. 5. A naive implementation of such a scheme is costly, as it requires multiple tree updates for every edge. However, Kolmogorov [7] showed that there exists an efficient implementation (which he called TRW-S) of such a scheme in the MAP case. This implementation may only be applied if the trees are monotonic chains, defined as follows: given an ordering of the nodes in a graph, a set of chains is monotonic if the order of nodes in the chain respects the given ordering. This structure allows one to reuse messages in a way that simultaneously implements operations on multiple trees. The scheduling of messages is important for guaranteeing convergence in this case. It turns out that one needs to scan nodes along the pre-specified order, first forward and then backward.

In the marginals case, the TRW algorithm of Wainwright [12] corresponds to optimizing over tree regions but is not provably convergent. In the next section we show how to derive a convergent algorithm for this case using our formalism.

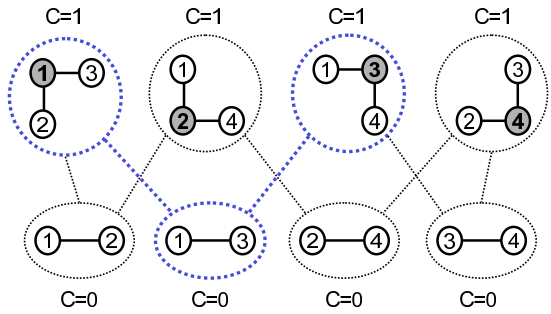

Figure 4: MPLP for pairs and singletons is equivalent to Heskes' algorithm with stars and pairs.

---

**Algorithm 4** The max product linear programming (MPLP) algorithm

Iterate over pairs of neighbouring nodes $<ij>$:

1. Set the message from $i$ to $<ij>$:
$$m_{i \to ij}^{t+1}(x_i) = \prod_{k \in N(i) \setminus j} m_{ik \to i}^t(x_i)$$
and equivalently from $j$ to $<ij>$

2. Update the pairwise beliefs of $<ij>$:
$$b_{ij}^{t+1}(x_i, x_j) \propto \sqrt{\Psi_{ij}(x_i, x_j) \cdot m_{i \to ij}^{t+1}(x_i) \cdot m_{j \to ij}^{t+1}(x_j)}$$

3. Set the messages from $<ij>$ to $i$ (and equivalently from $<ij>$ to $j$):
$$m_{ij \to i}^{t+1}(x_i) \propto \sqrt{\frac{\max_{x_j} \left( \Psi_{ij}(x_i, x_j) \cdot m_{j \to ij}^{t+1}(x_j) \right)}{m_{i \to ij}^{t+1}(x_i)}}$$

4. Set the beliefs of $i$ (and same for $j$):
$$b_i^{t+1}(x_i) \propto m_{ij \to i}^{t+1}(x_i) \cdot \prod_{k \in N(i) \setminus j} m_{ik \to i}^t(x_i)$$

---

## 5 New bound minimizing algorithms

By replacing the max with a sum (or vice versa) in the algorithms discussed in the previous section, we obtain algorithms that enforce a different type of consistency, and keep the same reparameterization and region graph as shown in the figures. Thus, the max-product version of Heskes' algorithm and the sum-product versions of TRW-S, MPLP and MSD are convergent with respect to the relevant bound.

The TRW-S sum-product case is especially interesting. In this case the relevant bound becomes the tree-reweighted log-partition function bound introduced in [12]. The message passing algorithm suggested in [12] does not generally converge. In contrast, the TRW-S sum-product algorithm is guaranteed to converge, since it is an instance of a TCBO algorithm for the sum case. Furthermore, this TRW-S variant differs from the algorithm in [12] only in the scheduling of messages.

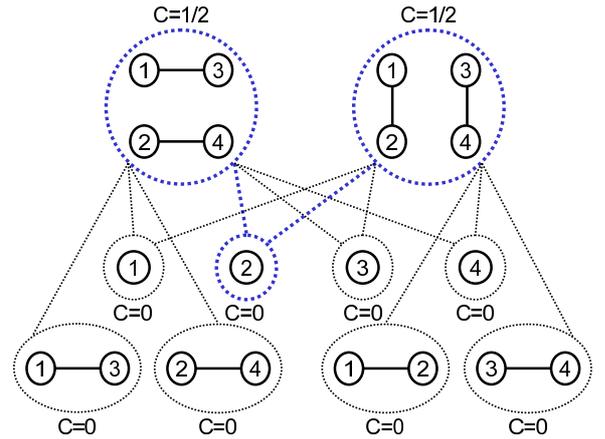

Figure 5: A region graph with chains, their pairwise and singleton components. Such graphs are used by the TRW-S algorithms. In the above example there are two chains, which are also monotonic chains since they agree with the node ordering $\{1, 2, 3, 4\}$. TRW-S may be viewed as a TCBO algorithm on the subgraph shown in the blue dashed line (and an additional subgraph corresponding to a pairwise component and all the chains that contain it).

An additional algorithm that can be easily shown to be convergent is two way GBP [14] with all double counting numbers $c_\alpha = 1$, both in the sum and in the max versions. At each iteration, two-way GBP updates only the beliefs of a region and one of its sub-regions, which is trivially a tree. The fact that it maintains reparameterization and enforces consistency can be shown directly. In fact, it can be shown that two way GBP with $c_\alpha = 1$ is identical to MSD.

## 6 Experiments

We present two experiments to illustrate the convergence of our sum and max algorithms. All algorithms were applied to an instance of a 10x10 "spin glass" with pairwise terms drawn randomly from $[-9, 9]$ and field from $[-1, 1]$. In each case we tested the new TCBO algorithms.

For estimating the log-partition, we ran TRW and considered a uniform distribution over 2 spanning forests in the graph: all horizontal and all vertical chains. These chains are monotone with respect to the node ordering $\{1, 2, ..., 100\}$. We ran TRW-S by following the nodes order first forward and then backward, updating each time only the messages in the direction of

**Algorithm 5** The sequential tree reweighted BP (TRW-S) algorithm

1. Iterate over edges $i \to j$ in a certain updating order, and set the message from $i$ to $j$:

$$m_{i \to j}(x_j) \propto \max_{x_i} \Psi_i(x_i) \Psi_{ij}^{1/\rho_{ij}}(x_i, x_j) \frac{\prod_{k \in N(i) \setminus j} m_{k \to i}^{\rho_{ik}}(x_i)}{m_{j \to i}^{1-\rho_{ij}}(x_i)}$$

2. After each iteration over all edges, update all singleton and pairwise beliefs:

$$b_i(x_i) \propto \Psi_i(x_i) \prod_{j \in N(i)} m_{j \to i}^{\rho_{ij}}(x_i)$$

$$b_{ij}(x_i, x_j) \propto \Psi_i(x_i) \Psi_j(x_j) \Psi_{ij}^{1/\rho_{ij}}(x_i, x_j)$$

$$\cdot \frac{\prod_{k \in N(i) \setminus j} m_{k \to i}^{\rho_{ik}}(x_i)}{m_{j \to i}^{1-\rho_{ij}}(x_i)} \frac{\prod_{k \in N(j) \setminus i} m_{k \to j}^{\rho_{jk}}(x_j)}{m_{i \to j}^{1-\rho_{ij}}(x_j)}$$

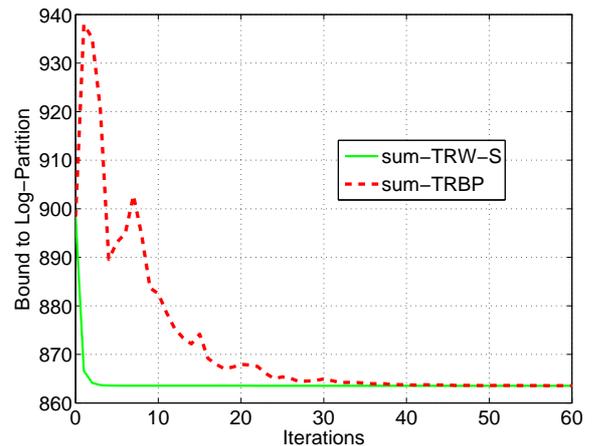

Figure 6: The bound on the log-partition in a 10x10 "spin-glass", obtained by sum-product TRW and TRW-S with edge probability appearances of 1/2. Note that the two algorithms differ only in the order of updates they perform. TRW-S follows the node ordering $\{1, 2, ..., 100\}$ that agrees with the monotonic chains, first forward and then backward. In the TRW implementation we followed the same nodes order in a forward manner.

the scan. Fig. 6 shows a comparison of this schedule to TRW where the node ordering is followed in a forward manner, and all outgoing messages are updated from each node. Both schedules keep a re-parameterization and provide a bound to the log-partition, yet only TRW-S monotonically decreases it at each iteration.

For the MAP case, we ran the max-product version of Heskes' algorithm using a region graph of pairs (with double counting numbers 1) and singletons (with double counting numbers 0). We also ran MPLP and MSD on the same problem. Fig. 7 shows the bounds obtained after each iteration. As can be seen, all three algorithms monotonically converged to the same value.

## 7 Discussion

Despite the empirical success of max-product and sum-product algorithms in applications, the original algorithms are not guaranteed to converge. Much research in recent years has therefore been devoted to devising convergent algorithms. Typically these recent algorithms are either max-product or sum-product and their proof of convergence is specific to the algorithm. Here we have presented a unified framework for convergent message passing algorithms and showed the importance of enforcing consistency in both sum-product and max-product algorithms. Not only does this analogy allow us to give a unified derivation for existing algorithms, it also gives an easy way to derive novel algorithms from existing ones by exchanging maximizations and summations.

Although many convergent algorithms are instances of our framework, it is worth pointing out two convergent algorithms that are not. The first is Hazan and Shashua's recent algorithm [3, 4] which works for provably convex double counting numbers (not necessarily positive as we are assuming). The second is ordinary BP on a single cycle, which can be shown to be convergent in both its sum and max product forms. We emphasize that even negative counting numbers can be handled by us in some cases, by using larger regions.

All the algorithms we discussed here in fact only updated star graphs in the region graph. Our conditions for monotonicity apply to general tree updates. However, it seems less straightforward to obtain general (non-star) tree updates that achieve (max or sum) consistency and reparameterization simultaneously. Interestingly, the tree based updates in [9] do monotonically decrease an upper bound but seem not to satisfy max-consistency. Thus, it remains an interesting challenge to find general tree updates that satisfy the consistency constraints, as these could be easily used interchangeably for MAP and marginals.

Perhaps the most intriguing result of our analysis is the importance of update schedule for obtaining convergence – a non-convergent algorithm becomes convergent when the right update schedule is used. It will be interesting to see whether convergent update schedules can be derived for an even larger class of message-passing algorithms.

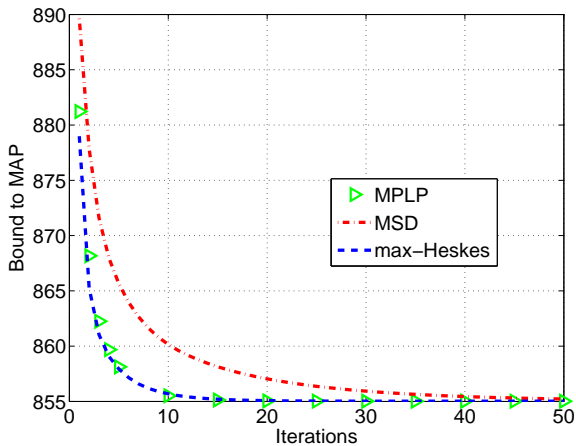

Figure 7: The bound on the MAP value in a 10x10 "spin-glass", obtained by the max-product version of Heskes' algorithm with double counting numbers 1 for regions of pairs and 0 for regions of singletons, and by MPLP and MSD.

## Appendix A  MPLP as a TCBO algorithm

MPLP can be derived as an instance of the max-product version to Heskes' algorithm, and thus to be shown as a TCBO. We derive it for the case where the potentials in the original graph are defined for pairs, and then the generalization to larger cliques is simple.

We consider the max-product version to Heskes' algorithm, applied to the region graph in Fig. 4. The region graph consists on a layer of all the stars $Si$ in the original graph. Each star $Si$ corresponds to a node $i$ and all its neighbours in the original graph, with the edges between them. The layer of intersection regions would be all the edges $<ij>$. The counting numbers would be $c_{Si} = 1$ for a star region and $c_{ij} = 0$ for an edge region. Assuming that the original potentials are $\theta_{ij}(x_i, x_j)$ for all pairs, we define the new potentials to be $\tilde{\theta}(x_i, x_j) = 0$ for the subsets, and $\tilde{\theta}(x_{Si}) = \frac{1}{2}\sum_{j \in N_i} \theta_{ij}(x_i, x_j)$ for the stars.

The messages passed in MPLP $m_{ij \to i}(x_i)$ are obtained from the messages $\mu_{ij \to Si}(x_i, x_j)$ passed in Heskes' algorithm through the transformation:

$$m^t_{ij \to i}(x_i) = \max_{x_j} \left( \frac{1}{2}\theta_{ij}(x_i, x_j) + \mu^t_{ij \to Si}(x_i, x_j) \right)$$

It is easy to verify that this transformation yields the same update rule and beliefs of MPLP, and the same bound is minimized.

In order to generalize our derivation for clusters larger than pairs, we should replace the edges and stars by clusters and "hyper stars" respectively, in which the centers are the intersections of the clusters.